\documentclass[10pt,twocolumn,letterpaper]{article}

\usepackage[pagenumbers]{cvpr} 

\definecolor{cvprblue}{rgb}{0.21,0.49,0.74}
\usepackage[pagebackref,breaklinks,colorlinks,allcolors=cvprblue]{hyperref}
\usepackage{pifont}
\usepackage[table]{xcolor}
\usepackage[most]{tcolorbox}
\usepackage{multirow,array,threeparttable}
\usepackage{graphicx,arydshln,xfrac,subcaption}
\usepackage{amsthm}

\def\confName{CVPR}
\def\confYear{2027}

\newtheorem{definition}{Definition}[section]

\newtheorem{observation}{Observation}[section]
\newcommand{\std}[1]{\,{\scriptsize$\pm$#1}}
\title{Disentangling Multimodal Rotary Positional Embedding in Vision-Language Models}

\author{Zhaoyi Liu\textsuperscript{1}, Haichuan Zhang\textsuperscript{2}, Xiaochuan Gong\textsuperscript{1}, Yancheng Zhu\textsuperscript{1} \\ Bowei Tian\textsuperscript{1}, Wanghao Ye\textsuperscript{1}, Yufan Zhang\textsuperscript{1}, Ang Li\textsuperscript{1}\\
\textsuperscript{1}University of Maryland, College Park\\
\textsuperscript{2}University of Utah\\
}

\begin{document}
\maketitle

\begin{abstract}
Vision-language models widely adopt Multimodal Rotary Position Embedding (MRoPE), an extension of Rotary Position Embedding (RoPE) that incorporates temporal, height, and width axes to encode the spatial structure of visual tokens. However, MRoPE also assigns height and width coordinates to text tokens that lack inherent spatial structure, introducing spatial positional effects into text-vision attention. Existing MRoPE variants explore different frequency allocations and coordinate layouts, with their designs suggesting different intuitions about these cross-modal effects: they may provide useful spatial references for associating language with visual regions, or introduce unnecessary positional preferences that hinder text-vision interactions. In this work, we disentangle intra-visual spatial structure from cross-modal spatial effects to investigate this question. We identify a constraint of shared coordinate layouts that prevents eliminating cross-modal spatial effects while preserving intra-visual spatial encoding, and develop controlled experiments to isolate their respective contributions. Our results across multiple model scales, MRoPE variants, and benchmarks show that cross-modal spatial effects provide little additional benefit as spatial references and can even impair model performance, while intra-visual spatial encoding remains useful for fine-grained grounding.
\end{abstract}    
\section{Introduction}
\label{sec:introduction}
\begin{figure}[ht]
    \centering
    \includegraphics[width=0.8\columnwidth]{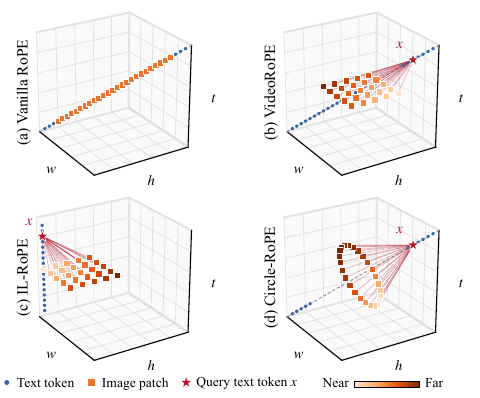}
    \definecolor{text-token}{HTML}{3567A8}
    \definecolor{query-token}{HTML}{B2182B}
    \caption{
    Illustration of different coordinate layouts for vanilla-RoPE
    and representative MRoPE variants in the temporal-height-width
    ($t,h,w$) space. Blue dots (\textcolor{text-token}{$\bullet$}) represent text tokens, orange squares (\textcolor{orange}{$\blacksquare$}) represent visual patches within an image, and the red star (\textcolor{query-token}{$\bigstar$}) marks a query text token $x$. The orange color gradient indicates the relative spatial proximity of visual patches to $x$ along the height and width axes, with lighter colors indicating nearer positions and darker colors indicating farther positions.
    }
    \label{fig:highlight}
\end{figure}
Rotary Position Embedding (RoPE)~\cite{Su2021RoFormerET} has achieved widespread success in large language models by applying position-dependent rotations to query and key vectors, allowing attention scores to encode relative positional information. One adoption in vision-language models (VLMs), hereafter referred to as vanilla RoPE, assigns one-dimensional position indices to a flattened sequence of text and visual tokens. Multimodal Rotary Position Embedding (MRoPE)~\cite{Wang2024Qwen2VLEV} extends this formulation by introducing three positional axes: temporal, height, and width. Specifically, it partitions the rotary dimensions into three groups, each rotated according to the corresponding positional coordinate. In this three-axes coordinate space, vanilla RoPE corresponds to a diagonal layout where all three coordinates share the same position index, as shown in Fig.~\ref{fig:highlight}(a). In contrast, MRoPE and its variants VideoRoPE~\cite{Wei2025VideoRoPEWM}(Fig.~\ref{fig:highlight}(b)) and IL-RoPE~\cite{Liao2025MogaoAO}(Fig.~\ref{fig:highlight}(c)) assign visual tokens height and width coordinates according to their positions in the image grid, explicitly encoding their two-dimensional spatial relationships. 

By placing text and visual tokens in the same coordinate space, these MRoPE methods thus also establish relative spatial relationships across modalities, giving rise to the intuition that text coordinates can serve as spatial references for associating language with visual regions.

However, these spatial references also introduce position-dependent effects into text-vision attention. Since RoPE encodes relative positions, the relative rotation between a text token and a visual patch depends on their height and width offsets as well. As illustrated by the varying spatial proximity of visual patches to the query text token in Fig.~\ref{fig:highlight}(b),  and (c), visual patches within a single image at different spatial locations have different relative height and width offsets from a given text token, regardless of whether text coordinates follow the diagonal or remain zero. Consequently, the spatial location of a visual patch can influence its attention score with a text token, even when their content representations are held fixed.

Circle-RoPE~\cite{Wang2025CircleRoPECD}, in contrast, seeks to eliminate this spatial dependence through a different coordinate layout. As illustrated in Fig.~\ref{fig:highlight}(d), it projects visual coordinates onto the annulus orthogonal to the diagonal text trajectory, aiming to equalize their Euclidean distances from each text token. These contrasting design motivations reveal a fundamental tension: cross-modal spatial dependencies may provide useful spatial references for associating language with visual regions, or they may introduce unnecessary spatial preferences that interfere with text-vision attention, as Circle-RoPE seeks to address. We highlight that this tension stems from the lack of inherent height and width coordinates for text tokens. However, Circle-RoPE's projection also changes the intra-visual relative positions among visual tokens. Moreover, as illustrated by the varying spatial proximity in Fig.~\ref{fig:highlight}(d), equal Euclidean distances do not necessarily imply identical relative RoPE rotations (see \S\ref{sec:method}). Thus, such geometric adjustments do not independently isolate cross-modal spatial effects from intra-visual spatial encoding, leaving a fundamental question unanswered:

\textit{Do these cross-modal spatial effects facilitate text-vision interactions by providing useful spatial references, or do they introduce unnecessary positional preferences that impair model performance?}

In this work, we empirically investigate this question by disentangling the effects of intra-visual spatial structure and cross-modal spatial dependencies in MRoPE. We first identify a constraint of shared coordinate layouts that prevents these two effects from being independently controlled. Building on this analysis, we use attention-specific coordinate assignments to construct controlled experiments to independently examine how these two components affect
model performance. Our main contributions are summarized as follows:
\begin{itemize}
\item We uncover an underexplored coupling between cross-modal spatial effects and intra-visual spatial encoding under shared coordinate layouts, and use this observation to construct controlled comparisons that isolate their respective contributions.
\item We conduct training-free evaluations on pretrained VLMs and controlled training experiments across model scales, MRoPE variants, and benchmarks to assess their effects on multimodal understanding and visual grounding.
\item We show that intra-visual spatial encoding benefits fine-grained visual grounding, whereas cross-modal spatial effects provide little additional benefit and can impair model performance.
\end{itemize}
\begin{figure*}[t]
    \centering
    \includegraphics[width=0.9\linewidth]{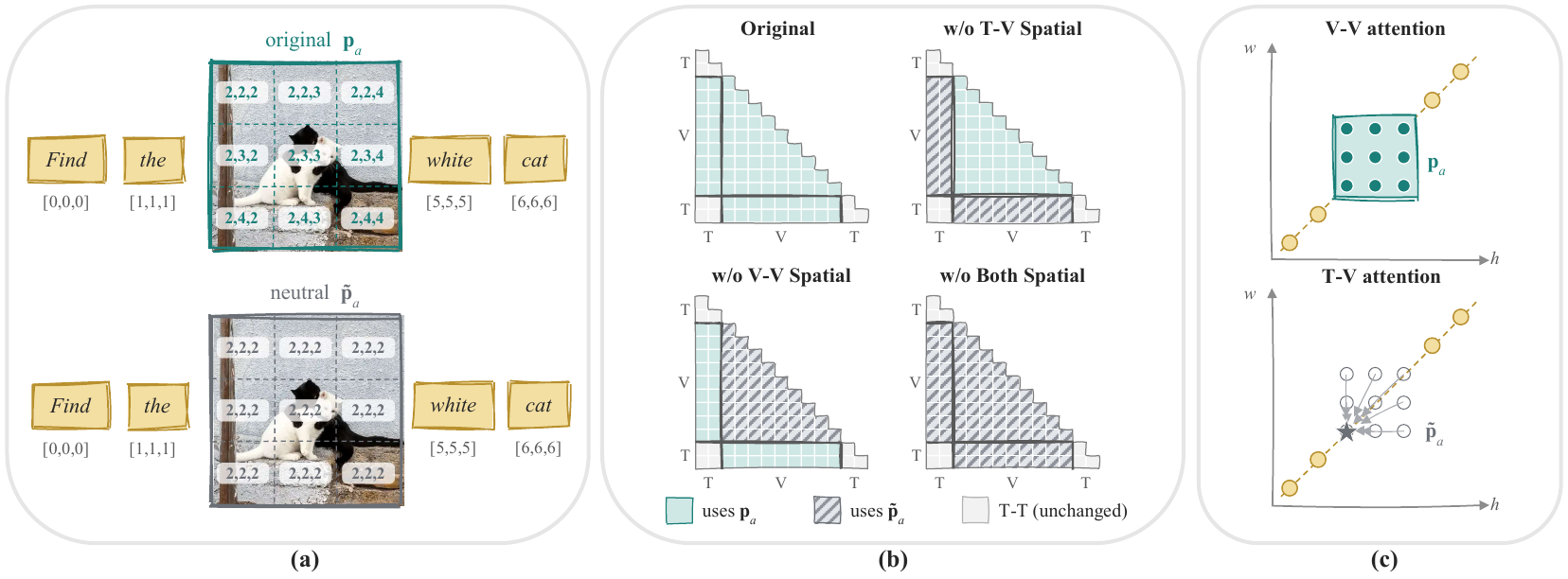}
    \caption{\textbf{Overview of attention-specific coordinate assignments for disentangling cross-modal and intra-visual spatial effects.} \textbf{(a) Two coordinate assignments for visual patches.} The original coordinates $\mathbf{p}_a=(t_{\mathrm{img}},h_a,w_a)$ preserve each patch's spatial location, while the neutral coordinates $\tilde{\mathbf{p}}_a=(t_{\mathrm{img}},\tilde{h},\tilde{w})$ assign identical height and width coordinates to all patches within an image. Here, $\tilde{h}=\tilde{w}=t_{\mathrm{img}}$ for the diagonal text layout. \textbf{(b) Attention-specific coordinate assignments.} Independently selecting original or neutral coordinates for text-vision (T-V) and vision-vision (V-V) attention yields four controlled configurations, while text coordinates and text-text (T-T) attention remain unchanged. \textbf{(c) Geometric view on the $\mathbf{(h,w)}$ plane.} In the w/o T-V Spatial configuration, original coordinates preserve spatial relationships among visual patches in V-V attention, while neutral coordinates yield identical relative rotations between any text token and all visual patches in T-V attention. This enables independent evaluation of the two spatial effects.}
    \label{fig:overview}
    \vspace{-1em}
\end{figure*}
\section{Preliminaries and Related Work}
\label{sec:preliminary}
\paragraph{Rotary Position Embedding (RoPE).}
RoPE~\cite{Su2021RoFormerET} encodes relative positional information in attention through position-dependent rotations of query and key vectors. For a flattened token sequence, let $m_i \in \mathbb{Z}$ denote the one-dimensional position index of token $i$, and let $\mathbf{q}_i,\mathbf{k}_i \in \mathbb{R}^d$ denote its query and key vectors, where $d$ is the rotary dimension of an attention head. RoPE partitions the rotary dimensions into $d/2$ two-dimensional blocks and applies a position-dependent rotation to each block:
\begin{equation}
    R(m)
    = \bigoplus_{\ell=1}^{d/2}
    \rho(\theta_\ell m),
    \quad
    \rho(\phi)=
    \begin{pmatrix}
        \cos\phi & -\sin\phi \\
        \sin\phi & \cos\phi
    \end{pmatrix},
\end{equation}
where $\oplus$ denotes block-diagonal concatenation and $\theta_\ell$ denotes the rotary frequency of block $\ell$. The resulting attention score between tokens $i$ and $j$, omitting the scaling factor, is
\begin{equation}
    s_{ij}
    = \left(R(m_i)\mathbf{q}_i\right)^\top
      \left(R(m_j)\mathbf{k}_j\right)
    = \mathbf{q}_i^\top
      R(m_j-m_i)\mathbf{k}_j.
\end{equation}
Thus, the positional contribution to attention depends on the relative position $m_j-m_i$ along the 1D token sequence.

\paragraph{Multimodal Rotary Position Embedding (MRoPE).}

Introduced in Qwen2-VL~\cite{Wang2024Qwen2VLEV}, MRoPE extends vanilla RoPE by replacing the scalar position index $m_i$ with a three-dimensional coordinate $\mathbf{p}_i=(t_i,h_i,w_i)$, representing the temporal, height , and width positions of token $i$. For visual tokens within the same image, the temporal coordinate is shared, while the height and width coordinates specify their locations in the 2D image grid. Specifically, let $a$ index a visual patch within an image. Its coordinates can be written as $\mathbf{p}_a=(t_{\mathrm{img}},h_a,w_a)$, where $t_{\mathrm{img}}$ is shared by all patches in the image, and $(h_a,w_a)$ denotes the patch's spatial location, as illustrated in Fig.~\ref{fig:highlight}. 

To incorporate these axes, MRoPE assigns each rotary block $\ell$ to a positional axis $a(\ell)\in\{t,h,w\}$.
The rotation operator becomes
\begin{equation}
    R(\mathbf{p})
    = \bigoplus_{\ell=1}^{d/2}
    \rho\left(\theta_\ell p_{a(\ell)}\right),
\end{equation}
where $p_{a(\ell)}$ denotes the coordinate of $\mathbf{p}$ along axis $a(\ell)$. Consequently, the attention score under MRoPE is
\begin{equation}
\begin{aligned}
    s_{ij}
    &= (R(\mathbf{p}_i)\mathbf{q}_i)^\top
       (R(\mathbf{p}_j)\mathbf{k}_j)= \mathbf{q}_i^\top
       R(\mathbf{p}_j-\mathbf{p}_i)\mathbf{k}_j.
\end{aligned}
\label{eq:mrpe_attention}
\end{equation}

For two visual patches $a$ and $b$ within the same image, their relative position is
\begin{equation}
    \mathbf{p}_b-\mathbf{p}_a
    = (0,h_b-h_a,w_b-w_a).
\end{equation}
Thus, MRoPE explicitly encodes the two-dimensional spatial relationships among visual tokens through their relative height and width coordinates.

\paragraph{Evolution of MRoPE.}
Existing MRoPE variants primarily explore two components of the above formulation: frequency allocation, which determines the mapping $a(\ell)$ from rotary blocks to positional axes, and coordinate layout, which specifies the position $\mathbf{p}$ assigned to each token.

For frequency allocation, VideoRoPE~\cite{Wei2025VideoRoPEWM} and HoPE~\cite{Li2025HoPEHO} prioritize lower frequencies for temporal encoding to improve long-range modeling. MRoPE-I~\cite{Huang2025RevisitingMP} instead interleaves the three axes across rotary dimensions, while MHRoPE~\cite{Huang2025RevisitingMP} distributes positional axes across different attention heads. MM-RoPE, proposed in the unified video generation model~\cite{Yuan2025Lumos1OA} adjusts the frequency allocation as well. These approaches primarily refine how positional information is represented across the rotary frequency spectrum to improve the performance beyond the original MRoPE.

For coordinate layout, existing methods explore different
assignments of $\mathbf{p}$ across modalities. MRoPE-I and VideoRoPE preserve compatibility with pretrained language models by assigning text tokens a diagonal layout. VideoRoPE further adjusts visual coordinates to improve spatial symmetry across modalities. In contrast, IL-RoPE~\cite{Liao2025MogaoAO} and Omni-RoPE~\cite{Wu2025OmniGen2ET} assign height and width coordinates of text tokens to be zero. Circle-RoPE~\cite{Wang2025CircleRoPECD} takes a different approach by projecting coordinates of vision tokens onto the annulus orthogonal to the diagonal text trajectory, aiming to equalize their Euclidean distances from text tokens and reduce cross-modal
spatial dependencies.

While these variants explore different frequency assignments $a(\ell)$ and coordinate layouts $\mathbf{p}$, their positional designs reflect contrasting approaches to cross-modal spatial effects. Diagonal and zero-spatial text layouts retain spatial dependencies in text-vision attention, whereas Circle-RoPE attempts to eliminate them. This contrast reflects a fundamental tension: cross-modal spatial effects may provide useful spatial references for associating language with visual regions, or introduce unnecessary positional preferences that hinder text-vision interactions. Whether these effects benefit or impair model performance, independently of intra-visual spatial structure, remains unexplored.
\section{Disentangling Cross-Modal Spatial Effects}
\label{sec:method}
To investigate the role of cross-modal spatial effects independently of intra-visual spatial structure, we first establish an operator-level criterion for spatial neutrality in text-vision attention. We then distinguish geometric equidistance from spatial neutrality and identify the inherent coupling between cross-modal and intra-visual spatial effects under shared coordinate layouts. Based on this analysis and observation, we use attention-specific coordinate assignments to independently examine their contributions through controlled experiments.

\subsection{Cross-Modal Spatial Neutrality}
\label{sec:defination}

As established in \S\ref{sec:preliminary}, MRoPE modulates attention through the relative positional rotation $R(\mathbf{p}_j-\mathbf{p}_i)$. Consider a text token $x$ at $\mathbf{p}_x=(t_x,h_x,w_x)$ and two visual patches $a,b$
within the same image, located at $\mathbf{p}_a=(t_{\mathrm{img}},h_a,w_a)$ and $\mathbf{p}_b=(t_{\mathrm{img}},h_b,w_b)$.
Since all patches in the image share the same temporal coordinate $t_{\mathrm{img}}$, their relative temporal offsets from the text token are identical. However, their relative height and width offsets vary with their spatial locations, introducing different positional rotations into text-vision attention.

\textbf{Intuition.} Our goal is to eliminate these spatial differences without removing content-dependent distinctions among visual patches. In particular, different patches may naturally receive different attention scores due to their visual content. What we seek to eliminate is the additional positional preference that arises solely from their spatial locations. Therefore, if two visual patches within a single image have identical content representations, their positional modulation with the same text token should be identical, regardless of where they are located within the image. This motivates the following definition of cross-modal spatial neutrality.

\begin{definition}[Cross-Modal Spatial Neutrality]
\label{eq:definition}

For a given text token $x$ and visual patches within the same image, text-vision attention is spatially neutral if, for any two patches $a,b$, their positional modulations are identical for arbitrary content vectors $\mathbf{q},\mathbf{k}\in\mathbb{R}^d$:
\begin{equation}
    \mathbf{q}^{\top}
    R(\mathbf{p}_a-\mathbf{p}_x)\mathbf{k}
    =
    \mathbf{q}^{\top}
    R(\mathbf{p}_b-\mathbf{p}_x)\mathbf{k},\quad
    \forall\,\mathbf{q},\mathbf{k},a,b.
    \label{eq:spatial_neutrality}
\end{equation}
\end{definition}

Requiring this equality for arbitrary content vectors ensures that spatial neutrality is a property of the positional encoding itself, rather than an accidental cancellation for particular content vectors. Since Eq.~\ref{eq:spatial_neutrality} holds for arbitrary content vectors, it is equivalent to the following operator-level condition (\textit{Proof in Appendix~\ref{app:neutrality_equivalence}}):
\begin{equation}
    R(\mathbf{p}_a-\mathbf{p}_x)
    =
    R(\mathbf{p}_b-\mathbf{p}_x),
    \quad \forall\,a,b.
    \label{eq:operator_neutrality}
\end{equation}
Thus, cross-modal spatial neutrality requires that all visual patches within an image induce identical relative RoPE rotations with respect to a given text token. Importantly, this condition constrains only positional modulation, allowing attention to distinguish visual regions based on their content.

\subsection{Spatial Coupling in MRoPE}
\label{sec:spatial_disentanglement}

\paragraph{Geometric Distance and Spatial Neutrality.}
As a concrete example, we consider Circle-RoPE~\cite{Wang2025CircleRoPECD}, which seeks to reduce cross-modal spatial dependencies through an annulus coordinate layout. As illustrated in Fig.~\ref{fig:highlight}(d), Circle-RoPE projects coordinates of visual patches of an image onto the annulus around the diagonal text trajectory, aiming to equalize their Euclidean distances from text tokens. Let $\mathbf{p}_a'$ and $\mathbf{p}_b'$ denote the projected coordinates of two visual patches within the same image. Circle-RoPE seeks to satisfy
\begin{equation}
    \|\mathbf{p}_a'-\mathbf{p}_x\|_2
    =
    \|\mathbf{p}_b'-\mathbf{p}_x\|_2.
    \label{eq:circle_distance}
\end{equation}

However, equal Euclidean distances do not guarantee identical relative RoPE rotations. As established in \S\ref{sec:preliminary}, MRoPE applies axis-specific rotations, where each rotary block $\ell$ encodes the relative displacement along its assigned axis $a(\ell)$. Therefore, the operator equality required by Eq.~\ref{eq:operator_neutrality} holds if and only if (\textit{Proof in Appendix~\ref{app:circle_proof}}):
\begin{equation}
    \theta_\ell
    [\mathbf{p}_a'
    -\mathbf{p}_b']_{a(\ell)}
    \in 2\pi\mathbb{Z},
    \qquad \forall\,\ell.
    \label{eq:circle_phase}
\end{equation}

Unlike Euclidean distance equality, this condition requires the relative rotation phases to agree in every rotary block. Thus, geometric equidistance alone does not guarantee the operator-level spatial neutrality defined in Eq.~\ref{eq:operator_neutrality}.

\paragraph{Coupling Under Shared Coordinate Layouts.}
Beyond this distinction, Circle-RoPE's projection also changes the relative 2D grid spatial relationships among visual patches, affecting intra-visual spatial encoding. These observations raise a broader question: Can any shared coordinate layout achieve cross-modal spatial neutrality while preserving intra-visual spatial encoding?

Consider the same text token $x$ and two visual patches $a,b$. By the relative rotation property of RoPE, their positional rotations satisfy
\begin{equation}
    R(\mathbf{p}_b-\mathbf{p}_a)
    =
    R(\mathbf{p}_a-\mathbf{p}_x)^\top
    R(\mathbf{p}_b-\mathbf{p}_x).
    \label{eq:rotation_coupling}
\end{equation}
This identity connects the intra-visual relative rotation between two patches to their respective cross-modal rotations with the same text token. Combining it with the spatial neutrality condition in Eq.~\ref{eq:operator_neutrality} yields the following observation (\textit{Proof in Appendix~\ref{app:neutrality_theorem}}).

\begin{observation}[Coupling of Cross-Modal and Intra-Visual Spatial Effects]
\label{eq:theorem}

Under a shared coordinate layout, for any text token $x$ and visual patches $a,b$ within the same image, cross-modal spatial neutrality holds if and only if all intra-visual relative rotations are trivial:
\begin{equation}
\begin{aligned}
R(\mathbf{p}_a-\mathbf{p}_x)
=R(\mathbf{p}_b-\mathbf{p}_x),
\quad \forall\,a,b&.\\
\qquad\iff
R(\mathbf{p}_b-\mathbf{p}_a)=I,
\quad \forall\,a,b&.
\end{aligned}
\end{equation}
\end{observation}

Observation~\ref{eq:theorem} identifies an inherent coupling between cross-modal spatial effects and intra-visual spatial encoding under shared coordinate layouts. This coupling is independent of frequency allocation or coordinate layout, and prevents the two spatial effects from being independently controlled through shared coordinate assignments.

\begin{table}[t]
\centering
\caption{Controlled configurations based on attention-specific visual patch coordinates. T-V and V-V denote text-vision and vision-vision attention, respectively. Each entry specifies the visual patch coordinates used for the corresponding attention, while text token coordinates remain unchanged. $\mathbf{p}_a$ and $\widetilde{\mathbf{p}}_a$ denote the original and spatially neutral coordinates for visual patch tokens.}
\label{tab:configurations}
\resizebox{\columnwidth}{!}{
\begin{tabular}{llcc}
\toprule
\textbf{Configuration}
& \textbf{Notation}
& \textbf{T-V Attention}
& \textbf{V-V Attention} \\
\midrule
Original
& Original
& $\mathbf{p}_a$
& $\mathbf{p}_a$ \\

Cross-Modal Neutral
& w/o T-V Spatial
& $\widetilde{\mathbf{p}}_a$
& $\mathbf{p}_a$ \\

Intra-Visual Neutral
& w/o V-V Spatial
& $\mathbf{p}_a$
& $\widetilde{\mathbf{p}}_a$ \\

Fully Neutral
& w/o Both Spatial
& $\widetilde{\mathbf{p}}_a$
& $\widetilde{\mathbf{p}}_a$ \\
\bottomrule
\end{tabular}
}
\end{table}
\subsection{Attention-Specific Coordinate Assignments}
\label{sec:controlled_configurations}

Observation~\ref{eq:theorem} identifies an inherent coupling between cross-modal spatial effects and intra-visual spatial encoding under shared coordinate layouts. While this coupling follows from RoPE's relative rotation property, its implications for independently studying cross-modal spatial effects in MRoPE have received limited attention in prior work. As such, to examine their respective contributions, we use attention-specific coordinate assignments to construct controlled comparisons.

Specifically, we consider two coordinate assignments for each visual patch $a$ within an image: its original MRoPE coordinates $\mathbf{p}_a=(t_{\mathrm{img}},h_a,w_a)$, and a spatially neutral assignment
\begin{equation}
    \widetilde{\mathbf{p}}_a
    =
    (t_{\mathrm{img}},\widetilde{h},\widetilde{w}),
    \qquad \forall\,a,
    \label{eq:neutral_tv_coordinates}
\end{equation}
where $\widetilde{h}=\widetilde{w}=t_{\mathrm{img}}$ for diagonal text layouts, and $\widetilde{h}=\widetilde{w}=0$ for zero-spatial text layouts, depending on the MRoPE variant we study. We thus place all visual patches on the corresponding text coordinate trajectory at $t_{\mathrm{img}}$, ensuring that text-vision relative rotations follow the same positional scheme as text-text interactions. This eliminates patch-dependent spatial differences without introducing arbitrary cross-modal positional offsets, providing a clean setting for isolating cross-modal spatial effects. 

The original coordinates retain each patch's spatial location, whereas the neutral coordinates assign identical height and width positions to all patches within an image. By selecting between these two assignments for different attention computations, we can independently control their spatial positional effects. For text-vision attention, using $\widetilde{\mathbf{p}}_a$ gives
\begin{equation}
\begin{aligned}
    \widetilde{R}_{xa}
    &=
    R(\widetilde{\mathbf{p}}_a-\mathbf{p}_x)\\
    &=
    R(t_{\mathrm{img}}-t_x,\,
      \widetilde{h}-h_x,\,
      \widetilde{w}-w_x),
    \qquad \forall\,a.
\end{aligned}
\label{eq:neutral_tv_rotation}
\end{equation}
Since this rotation is identical across all patches within the image, it satisfies Definition~\ref{eq:definition}, eliminating patch-dependent cross-modal spatial effects. Similarly, using the neutral coordinates for intra-visual attention yields
\begin{equation}
    R(\widetilde{\mathbf{p}}_b-\widetilde{\mathbf{p}}_a)
    =R(0,0,0)=I,
    \qquad \forall\,a,b,
    \label{eq:neutral_vv_rotation}
\end{equation}
eliminating intra-visual relative positional rotations. Using the original coordinates instead preserves the corresponding MRoPE rotations.

By independently selecting between the original and neutral coordinates for text-vision and vision-vision attention, we construct four controlled evaluation configurations, as illustrated in Fig.~\ref{fig:overview} and summarized in Table~\ref{tab:configurations}. Comparing configurations that differ only in their text-vision coordinates isolates the contribution of cross-modal spatial effects, while varying the vision-vision coordinates reveals the role of intra-visual spatial encoding. Together, these comparisons allow us to examine whether cross-modal spatial effects provide useful spatial references or introduce unnecessary positional preferences, and how their contributions depend on intra-visual spatial encoding.

For brevity, we use the notations in Table~\ref{tab:configurations} to denote the four configurations throughout the following sections.
\section{Experiments}
\label{sec:experiments}
\subsection{Experimental Setup}
\label{sec:exp_setup}

\paragraph{Model Initialization.}
To investigate the contributions of cross-modal spatial effects, we conduct controlled training experiments using the four configurations in Table~\ref{tab:configurations}. By independently controlling cross-modal and intra-visual spatial encoding during training, we aim to determine whether cross-modal spatial effects facilitate model performance or introduce unnecessary positional preferences.

To this end, we construct models following the Qwen3-VL~\cite{Bai2025Qwen3VLTR} architecture at two scales, 2B and 4B, avoiding positional dependencies inherited from previous multimodal training. Specifically, we initialize the vision encoder from the original Qwen3-VL model and the language backbone from the corresponding pretrained Qwen3~\cite{qwen3technicalreport} base model, while randomly initializing the visual merger and vision-language projector. For each model scale, all four configurations share identical initial parameters and differ only in the visual coordinates assigned during attention computation, ensuring a controlled comparison of spatial positional effects.

\begin{table}[t]
    \centering
    \caption{
    Evaluation of official Qwen3-VL models under different attention-specific coordinate configurations across three model scales. General evaluates visual understanding and spatial understanding, where Visual and Spatial report the average scores across 10 visual understanding and 4 spatial understanding benchmarks, respectively. Grounding evaluates referring expression grounding, reporting accuracy at IoU thresholds of 0.5, 0.7, and 0.9, averaged across RefCOCO, RefCOCO+, and RefCOCOg. Higher scores indicate better performance.
    }
    \resizebox{\columnwidth}{!}{
    \begin{tabular}{lccccc}
    \toprule
    \multirow{2}{*}{\textbf{Configuration}}
    & \multicolumn{2}{c}{\textbf{General}}
    & \multicolumn{3}{c}{\textbf{Grounding}} \\
    \cmidrule(lr){2-3}
    \cmidrule(lr){4-6}
    & \textbf{Visual}
    & \textbf{Spatial}
    & \textbf{ACC@0.5}
    & \textbf{ACC@0.7}
    & \textbf{ACC@0.9} \\
    \midrule
    \multicolumn{6}{c}{\cellcolor[gray]{0.95}\textbf{\textit{Qwen3-VL-2B}}} \\
    \midrule
    Original
    &70.3 &43.6  &82.4 &74.0 &44.8 \\
    w/o T-V Spatial
    &\textbf{70.5} &\textbf{43.9}  &\textbf{82.5} &\textbf{74.3} &\textbf{45.0} \\
    w/o V-V Spatial
    &63.4 &42.1  &18.5 &6.3 &1.0 \\
    w/o Both Spatial
    &63.4 &42.3  &18.7 &6.3 &1.0 \\

    \midrule
    \multicolumn{6}{c}{\cellcolor[gray]{0.95}\textbf{\textit{Qwen3-VL-4B}}} \\
    \midrule
    Original
    &75.9 &49.4  &\textbf{87.1} &80.1 &\textbf{56.0} \\
    w/o T-V Spatial
    &\textbf{76.0} &\textbf{49.7}  &87.0 &80.0 &55.7 \\
    w/o V-V Spatial
    &68.9 &47.0  &16.3 &5.0 &0.8 \\
    w/o Both Spatial
    &68.6 &47.2  &16.4 &5.0 &0.7 \\

    \midrule
    \multicolumn{6}{c}{\cellcolor[gray]{0.95}\textbf{\textit{Qwen3-VL-8B}}} \\
    \midrule

    Original
    &\textbf{78.4} &49.7  &\textbf{86.1} &79.0 &\textbf{54.4} \\
    w/o T-V Spatial
    &78.1 &\textbf{50.1}  &\textbf{86.1} &\textbf{79.1} &54.3 \\
    w/o V-V Spatial
    &74.1 &48.6  &5.3 &1.9 &0.5 \\
    w/o Both Spatial
    &73.7 &48.4  &5.3 &1.9 &0.5 \\

    \bottomrule
    \end{tabular}
    }
    \label{tab:training_free}
\end{table}

\begin{table*}[t]
    \centering
    \caption{
    Controlled training results of Qwen3-VL-2B/4B
    under our four spatial configurations using MRoPE-I~\cite{Huang2025RevisitingMP}.
    All configurations follow the same training protocol.
    Results are reported as mean$\pm$std over three random seeds (42, 123, 3407).
    Visual and Spatial denote general visual understanding
    and spatial understanding benchmarks, respectively.
    The final rows report average scores across the
    10 visual and 4 spatial benchmarks.
    Higher values indicate better performance.
    }
    \resizebox{0.85\linewidth}{!}{
    \begin{tabular}{cl cccc|cccc}
    \toprule
    & \multirow{2}{*}{\textbf{Benchmark}} & \multicolumn{4}{c}{\textbf{Qwen3-VL-2B}} & \multicolumn{4}{c}{\textbf{Qwen3-VL-4B}} \\
    \cmidrule(lr){3-6} \cmidrule(lr){7-10}
    & & \multirow{2}{*}{Original} & w/o T-V & w/o V-V & w/o Both & \multirow{2}{*}{Original} & w/o T-V & w/o V-V & w/o Both \\
    & & & Spatial & Spatial & Spatial & & Spatial & Spatial & Spatial \\
    \midrule
    \multirow{10}{*}{\textbf{Visual}}
    & MMStar      & 44.5\std{0.1} & \underline{44.6}\std{0.0} & 44.2\std{0.3} & \textbf{45.2}\std{0.7} & 48.8\std{0.4} & \textbf{50.0}\std{1.1} & 48.8\std{0.8} & \underline{49.5}\std{0.1} \\
    & SEEDBench   & 70.9\std{0.3} & \textbf{71.2}\std{0.1} & \underline{71.1}\std{0.2} & 70.8\std{0.1} & 72.5\std{0.3} & 72.8\std{0.3} & \underline{72.9}\std{0.0} & \textbf{73.0}\std{0.1} \\
    & MMBench     & 68.9\std{0.4} & \textbf{69.4}\std{0.3} & \underline{69.2}\std{0.7} & 68.1\std{0.3} & 72.9\std{0.9} & \textbf{73.7}\std{1.2} & 73.4\std{0.3} & \underline{73.6}\std{0.2} \\
    & RealWorldQA & \underline{62.5}\std{0.8} & \textbf{63.0}\std{1.0} & 60.0\std{1.2} & 60.7\std{1.7} & 64.8\std{0.9} & \textbf{65.9}\std{0.6} & 63.3\std{0.8} & \underline{65.7}\std{0.3} \\
    & BLINK       & \underline{43.5}\std{0.2} & 43.4\std{0.9} & \textbf{44.4}\std{1.0} & 43.2\std{0.0} & \textbf{46.4}\std{1.0} & 46.2\std{0.7} & \underline{46.3}\std{0.2} & 46.2\std{0.2} \\
    & V*          & \textbf{66.0}\std{0.5} & 64.1\std{0.6} & \underline{64.7}\std{2.4} & 63.0\std{0.3} & \underline{65.3}\std{0.3} & \textbf{66.1}\std{2.1} & 62.0\std{0.6} & 62.1\std{0.6} \\
    & AI2D        & \textbf{72.1}\std{0.0} & 71.0\std{0.5} & 70.6\std{0.1} & \underline{71.5}\std{0.3} & 77.0\std{0.6} & \textbf{77.7}\std{0.4} & \underline{77.4}\std{0.4} & 77.1\std{0.1} \\
    & OCRBench    & \underline{58.0}\std{0.6} & 57.7\std{1.3} & \textbf{58.4}\std{0.2} & 57.4\std{0.5} & \textbf{59.1}\std{0.8} & \underline{58.5}\std{0.6} & 58.0\std{0.4} & 58.1\std{0.2} \\
    & DocVQA      & \underline{72.1}\std{0.0} & \textbf{72.3}\std{0.4} & 71.3\std{0.3} & 71.3\std{0.1} & \underline{75.5}\std{0.4} & \textbf{75.6}\std{0.4} & 74.1\std{0.1} & 73.7\std{0.1} \\
    & InfoVQA     & \underline{33.1}\std{0.6} & \textbf{34.4}\std{0.4} & 32.0\std{0.5} & 32.4\std{0.3} & \textbf{38.1}\std{1.1} & \underline{37.1}\std{0.7} & 35.6\std{0.4} & 36.4\std{0.2} \\
    \midrule
    \multirow{4}{*}{\textbf{Spatial}}
    & 3DSRBench   & \textbf{38.7}\std{1.6} & \underline{37.1}\std{0.3} & 36.7\std{0.5} & 36.7\std{1.3} & 35.6\std{1.8} & 38.0\std{0.5} & \textbf{39.2}\std{0.0} & \underline{38.8}\std{0.0} \\
    & Omni3DBench & 32.0\std{1.3} & \textbf{34.5}\std{3.0} & 30.9\std{1.1} & \underline{33.0}\std{1.7} & \underline{35.8}\std{0.8} & \textbf{36.3}\std{0.3} & 35.7\std{1.2} & 34.9\std{0.6} \\
    & Spatial457  & 35.7\std{0.4} & \textbf{36.3}\std{0.2} & 35.9\std{0.1} & \underline{36.1}\std{0.2} & \underline{36.4}\std{0.0} & 35.8\std{0.8} & \textbf{36.6}\std{0.2} & 36.3\std{0.1} \\
    & SpatialEval & 33.3\std{0.9} & \textbf{36.6}\std{2.1} & 34.9\std{0.1} & \underline{35.8}\std{1.6} & \underline{44.2}\std{2.0} & \textbf{47.2}\std{1.1} & 43.7\std{0.6} & 43.3\std{0.5} \\
    \midrule
    \multirow{2}{*}{\textbf{Overall}}
    & Visual  & \textbf{59.2}\std{0.2} & \underline{59.1}\std{0.3} & 58.6\std{0.3} & 58.4\std{0.2} & \underline{62.0}\std{0.1} & \textbf{62.4}\std{0.1} & 61.2\std{0.4} & 61.5\std{0.2} \\
    & Spatial & 34.9\std{0.4} & \textbf{36.1}\std{0.2} & 34.6\std{0.2} & \underline{35.4}\std{1.1} & 38.0\std{0.5} & \textbf{39.3}\std{0.2} & \underline{38.8}\std{0.4} & 38.3\std{0.3} \\
    \bottomrule
    \end{tabular}
    }
    \label{tab:training_mropei}
\end{table*}

\noindent \textbf{Training Protocol.}
We adopt a two-stage training procedure. In the first stage, we train the vision-language projector and visual merger on LLaVA-Pretrain-558K~\cite{Liu_2024_CVPR} while freezing the remaining components to establish visual-language alignment. In the second stage, we unfreeze the entire model and jointly fine-tune all components using LLaVA-NeXT-779K~\cite{liu2024llavanext} with additional grounding samples. All four configurations in Table~\ref{tab:configurations} follow the same training procedure, including datasets and optimization settings, to ensure fair comparisons.

\noindent\textbf{Evaluation Benchmarks.}
We evaluate model performance across three categories.
\begin{itemize}
    \item \textit{Visual Understanding.} We use MMStar~\cite{Chen2024AreWO}, SEEDBench~\cite{Li2023SEEDBenchBM}, MMBench~\cite{Liu2023MMBenchIY}, RealWorldQA~\cite{xai2024grok15v}, BLINK~\cite{Fu2024BLINKML}, V* Bench~\cite{Wu2023VGV}, AI2D~\cite{Kembhavi2016ADI}, OCRBench~\cite{Liu2023OCRBenchOT}, DocVQA~\cite{Mathew2020DocVQAAD}, and InfoVQA~\cite{Mathew2021InfographicVQA}, covering general visual perception, reasoning, and document understanding.
    \item \textit{Spatial Understanding.} We include 3DSRBench~\cite{Ma20243DSRBENCHAC}, Omni3DBench~\cite{Marsili2025VisualAA}, Spatial457~\cite{Wang2025Spatial457AD}, and SpatialEval~\cite{Wang2024IsAP} to assess spatial understanding and reasoning capabilities.
    \item \textit{Referring Expression Grounding.} We evaluate on the RefCOCO~\cite{Kazemzadeh2014ReferItGameRT} family, including RefCOCO, RefCOCO+, and RefCOCOg. These benchmarks assess fine-grained localization of visual regions based on natural language descriptions.
\end{itemize}
We use Lmms-Eval~\cite{zhang2024lmmsevalrealitycheckevaluation} for the RefCOCO family and VLMEvalKit~\cite{Duan2024VLMEvalKitAO} for all other benchmarks, both widely adopted evaluation frameworks, to ensure standardized and reliable performance assessment.

\noindent\textbf{Evaluation Metrics.}
For general visual and spatial understanding benchmarks, we report their standard evaluation metrics. For referring expression grounding, we evaluate the predicted bounding boxes using intersection-over-union (IoU) and localization accuracy at IoU thresholds of 0.5, 0.7, and 0.9, denoted as ACC@0.5, ACC@0.7, and ACC@0.9, respectively. All scores are reported as percentages.

\begin{figure*}[t]
    \centering
    \includegraphics[width=0.86\linewidth]{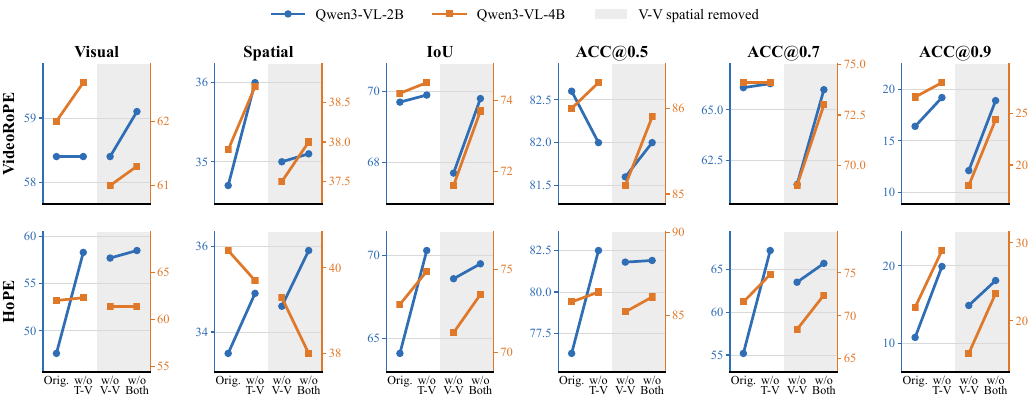}
    \caption{
    Controlled training results across two MRoPE variants
    (VideoRoPE~\cite{Wei2025VideoRoPEWM}, HoPE~\cite{Li2025HoPEHO}) under four spatial configurations (seed 42).
    Each row represents an MRoPE variant, while columns report
    average visual understanding, spatial understanding, and
    referring expression grounding performance.
    Blue circles and orange squares denote Qwen3-VL-2B and Qwen3-VL-4B,
    respectively, with their corresponding scores shown on the
    left and right y-axes.
    The shaded regions indicate configurations without
    intra-visual spatial encoding. \textit{More quantitative results on IL-RoPE~\cite{Liao2025MogaoAO} can be found in Appendix~\ref{app:ilrope}.}
    }
    \label{fig:transfer_mrope}
\end{figure*}
\begin{figure*}[t]
    \centering
    \begin{minipage}[t]{0.9\linewidth}
        \centering
        \includegraphics[width=\linewidth]{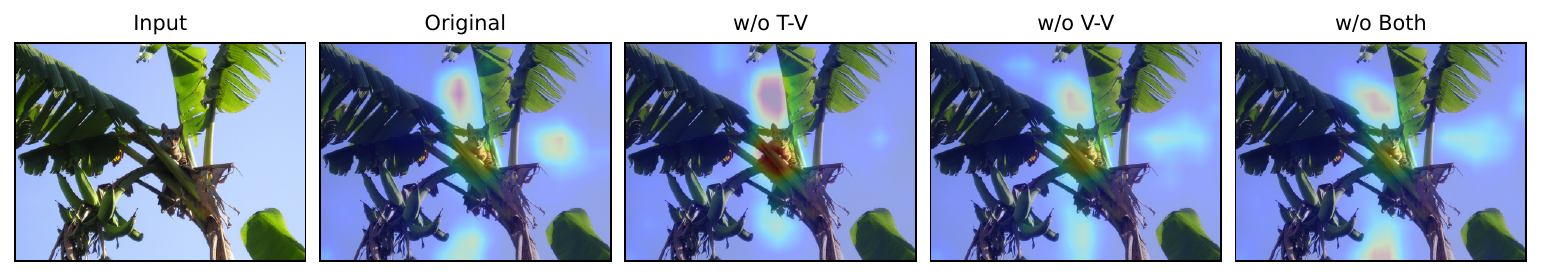}
    \end{minipage}
    \newline
    \begin{minipage}{0.9\linewidth}
        Where is the \textcolor{orange}{cat}?
    \end{minipage}

    \begin{minipage}[t]{0.9\linewidth}
        \centering
        \includegraphics[width=\linewidth]{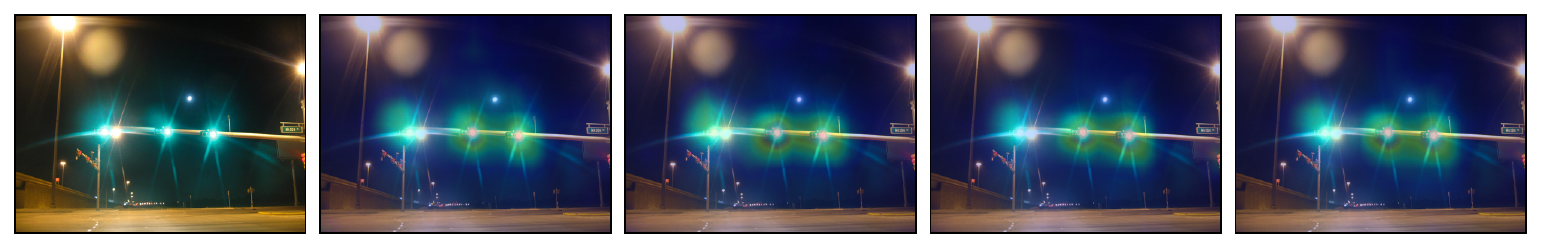}
    \end{minipage}
    \newline
    \begin{minipage}{0.9\linewidth}
        How many traffic lights are \textcolor{orange}{green}?
    \end{minipage}
    \caption{Visualization of attention maps for Qwen3-VL-4B~\cite{Bai2025Qwen3VLTR} (middle layers, 12-23)
    using the MRoPE-I~\cite{Huang2025RevisitingMP} under four spatial configurations.
    The orange-highlighted words indicate the query text token used to compute attention over visual patches. Warmer colors indicate higher relative attention intensity. \textit{More visualizations can be found in Appendix~\ref{app:attn}.}}
    \label{fig:visualization}
\end{figure*}
\textit{More experimental details can be found in Appendix~\ref{app:implementation_details}.}
\subsection{Probing Cross-Modal Spatial Effects in Pretrained Models}
\label{sec:pretrained_probe}
The official Qwen3-VL~\cite{Bai2025Qwen3VLTR} models are trained with MRoPE-I, which retains spatial positional differences among visual patches in text-vision attention. These cross-modal spatial effects are present throughout training. We therefore investigate whether MRoPE provides useful spatial references for associating language with visual regions by directly probing the officially released Qwen3-VL checkpoints.
\footnote{%
  \begin{tabular}{@{}l@{}}
    \url{https://huggingface.co/Qwen/Qwen3-VL-2B-Instruct} \\
    \url{https://huggingface.co/Qwen/Qwen3-VL-4B-Instruct} \\
    \url{https://huggingface.co/Qwen/Qwen3-VL-8B-Instruct}
  \end{tabular}%
}
Specifically, we evaluate the official Qwen3-VL-2B, 4B, and 8B checkpoints under the four configurations in Table~\ref{tab:configurations}. We apply our attention-specific coordinate assignments during inference without updating any model parameters, allowing us to assess the contribution of cross-modal and intra-visual spatial effects in pretrained models.

As shown in Table~\ref{tab:training_free}, removing cross-modal spatial effects has negligible impact across all three model scales. Compared with Original, w/o T-V Spatial changes the overall performance on general benchmarks by at most 0.4\%, with similarly minor differences in grounding performance. In contrast, removing intra-visual spatial encoding consistently degrades performance, particularly on referring expression grounding. For example, ACC@0.5 drops from 82.4\% to 18.5\%, 87.1\% to 16.3\%, and 86.1\% to 5.3\% for the 2B, 4B, and 8B models, respectively. Removing both spatial effects produces similarly severe degradation.

These consistent trends challenge the general intuition that cross-modal spatial effects provide useful spatial references for associating language with visual regions. Our findings suggest that pretrained models do not substantially rely on these effects, despite being trained with them. Nevertheless, this does not necessarily imply that imposing these positional preferences during training comes at no cost. Since MRoPE directly introduces these preferences into text-vision attention, models may have learned to adapt to their influence. This motivates further investigation into how the presence of these effects throughout training influences model performance.

\begin{table}[t]
    \centering
    \caption{
    Controlled training results on referring expression
    grounding using Qwen3-VL-2B/4B with MRoPE-I.
    IoU reports mean intersection-over-union, while
    ACC@0.5, ACC@0.7, and ACC@0.9 measure localization
    accuracy at the corresponding IoU thresholds.
    All metrics are aggregated over the evaluated splits
    of RefCOCO, RefCOCO+, and RefCOCOg and reported as
    percentages (mean$\pm$std over three random seeds: 42, 123, 3407).
    Higher values indicate better performance.
    }
    \resizebox{\columnwidth}{!}{
    \begin{tabular}{lcccc}
    \toprule
    \multirow{2}{*}{\textbf{Configuration}}
    & \multicolumn{4}{c}{\textbf{Grounding}} \\
    \cmidrule(lr){2-5}
    & \textbf{IoU}
    & \textbf{ACC@0.5}
    & \textbf{ACC@0.7}
    & \textbf{ACC@0.9} \\
    \midrule
    \multicolumn{5}{c}{\cellcolor[gray]{0.95}\textbf{\textit{Qwen3-VL-2B}}} \\
    \midrule
    Original
    & 70.1\std{0.3} & 82.5\std{0.4} & \underline{66.8}\std{0.5} & \underline{19.3}\std{0.3} \\
    w/o T-V Spatial
    & \textbf{70.7}\std{0.2} & \textbf{82.9}\std{0.4} & \textbf{67.6}\std{0.3} & \textbf{20.5}\std{0.4} \\
    w/o V-V Spatial
    & 69.8\std{0.2} & 82.2\std{0.4} & 66.4\std{0.3} & 18.6\std{0.6} \\
    w/o Both Spatial
    & \underline{70.3}\std{0.3} & \underline{82.8}\std{0.4} & \underline{66.8}\std{0.6} & 18.9\std{0.4} \\
    \midrule
    \multicolumn{5}{c}{\cellcolor[gray]{0.95}\textbf{\textit{Qwen3-VL-4B}}} \\
    \midrule
    Original
    & \underline{74.7}\std{0.4} & \underline{86.4}\std{0.5} & \underline{74.2}\std{0.5} & \underline{28.0}\std{0.7} \\
    w/o T-V Spatial
    & \textbf{75.0}\std{0.2} & \textbf{86.5}\std{0.3} & \textbf{74.8}\std{0.2} & \textbf{29.2}\std{0.3} \\
    w/o V-V Spatial
    & 73.6\std{0.2} & 85.9\std{0.3} & 72.4\std{0.4} & 23.9\std{0.2} \\
    w/o Both Spatial
    & 74.1\std{0.2} & 86.2\std{0.3} & 73.7\std{0.4} & 26.2\std{0.1} \\
    \bottomrule
    \end{tabular}
    }
    \label{tab:training_mropei_grounding}
\end{table}

\subsection{Controlled Analysis of Spatial Positional Effects}
\label{sec:controlled_training}

Building on the preceding probe, we further investigate whether cross-modal spatial effects introduce a performance cost by independently controlling cross-modal and intra-visual spatial effects throughout training. Specifically, we train Qwen3-VL-2B/4B under our four configurations from identical clean-start initializations, allowing us to isolate their respective contributions to model performance. We first present detailed results using MRoPE-I~\cite{Huang2025RevisitingMP} as in Tables~\ref{tab:training_mropei} and~\ref{tab:training_mropei_grounding}.

As shown, removing cross-modal spatial effects preserves general benchmark performance while improving fine-grained visual grounding across both model scales. As shown in Table~\ref{tab:training_mropei}, compared with Original, w/o T-V Spatial improves average spatial understanding by 1.2\%/1.3\% for Qwen3-VL-2B/4B, while achieving comparable visual understanding performance. For grounding (Table~\ref{tab:training_mropei_grounding}), ACC@0.9 increases by 1.2\%/1.2\%, respectively. A similar trend is observed even without intra-visual spatial encoding: w/o Both Spatial outperforms w/o V-V Spatial by 0.3\%/2.3\% in ACC@0.9. Together, these comparisons suggest that cross-modal spatial effects provide little additional benefit as spatial references and can introduce unnecessary positional preferences that impair fine-grained grounding performance.

This finding is particularly noteworthy when contrasted with intra-visual spatial encoding, which captures the actual spatial relationships among visual patches. Removing intra-visual spatial encoding from Original reduces grounding ACC@0.9 by 0.7\%/4.1\% for Qwen3-VL-2B/4B. Together, these results highlight a central tension in MRoPE: the same framework that benefits from modeling intra-visual spatial structure can also incur a performance cost from the cross-modal positional preferences it introduces.

Fig.~\ref{fig:visualization} visualizes the attention maps of the trained Qwen3-VL-4B under four controlled configurations. Removing cross-modal spatial effects produces more concentrated attention on task-relevant visual regions.

\subsection{Generalization Across MRoPE Variants}
\label{sec:rope_generalization}

The preceding controlled experiments show that cross-modal spatial effects can impair model performance under MRoPE-I. A natural question is whether this finding stems from its specific positional design or reflects a more general property of cross-modal spatial effects. To investigate this, we extend our experiments to VideoRoPE~\cite{Wei2025VideoRoPEWM} and HoPE~\cite{Li2025HoPEHO}, which differ in coordinate layouts and frequency assignments. We evaluate all four attention-specific configurations on Qwen3-VL-2B and 4B under the same training protocol.

As shown in Fig.~\ref{fig:transfer_mrope}, removing cross-modal spatial effects improves fine-grained grounding across all variants and model scales. Compared with Original, w/o T-V Spatial improves ACC@0.9 by 2.8\%/1.4\% for VideoRoPE and 9.1\%/7.3\% for HoPE on Qwen3-VL-2B/4B, respectively. Grounding IoU and ACC@0.7 also improve in most cases, while visual understanding is generally improved or maintained. Even without intra-visual spatial encoding, w/o Both Spatial outperforms w/o V-V Spatial by 3.2\%-7.7\% in ACC@0.9 for VideoRoPE and HoPE.

The gains from removing cross-modal spatial effects are also comparable to those from preserving intra-visual spatial encoding. Across the four model–variant combinations, removing cross-modal spatial effects yields an average ACC@0.9 gain of 5.2\% (w/o T-V Spatial vs. Original), compared with the 2.8\% gain from preserving intra-visual spatial encoding (w/o T-V Spatial vs. w/o Both Spatial). These consistent trends provide further evidence that cross-modal spatial effects can introduce detrimental positional preferences not depending on single MRoPE design.
\section{Conclusion and Discussion}
\label{sec:conclusion}
In this work, we disentangle intra-visual spatial structure from cross-modal spatial effects in MRoPE. Our findings reveal that cross-modal spatial effects provide little additional benefit as spatial references and can introduce unnecessary positional preferences that impair model performance. Together, our analysis and empirical findings contribute to a better understanding of the distinct roles of spatial positional information within and across modalities, which may help inform future multimodal positional encoding designs. \textit{Limitations are discussed in Appendix~\ref{app:limitation}.}

{
    \small
    \bibliographystyle{ieeenat_fullname}
    \bibliography{main}
}

\clearpage
\setcounter{page}{1}
\maketitlesupplementary


\section{More Quantitative Results}
\label{app:ilrope}
We further evaluate IL-RoPE~\cite{Liao2025MogaoAO}, which assigns zero height and width coordinates to text tokens, to examine the generality of our findings across different coordinate layouts. As shown in Table~\ref{tab:ilrope}, removing cross-modal spatial effects improves spatial understanding by 2.1\% and grounding ACC@0.9 by 1.5\%, while removing intra-visual spatial encoding reduces ACC@0.9 by 2.6\%.

\section{Limitations}
\label{app:limitation}
Our controlled training experiments demonstrate consistent trends across two model sizes (2B/4B) and multiple MRoPE variants. However, whether these findings generalize to the substantially larger training regimes used in large-scale VLMs remains to be investigated.

\begin{table}[t]
    \centering
    \caption{
    Controlled training results on Qwen3-VL-4B with ILRoPE~\cite{Liao2025MogaoAO} (seed 42). Visual and Spatial report the average scores across 10 visual understanding and 4 spatial understanding benchmarks, respectively.
    IoU reports mean intersection-over-union, while
    ACC@0.5, ACC@0.7, and ACC@0.9 measure localization
    accuracy at the corresponding IoU thresholds.
    All metrics are aggregated over the evaluated splits
    of RefCOCO, RefCOCO+, and RefCOCOg and reported as
    percentages. Higher values indicate better performance.
    }
    \resizebox{\columnwidth}{!}{
    \begin{tabular}{lcccccc}
    \toprule
    \multirow{2}{*}{\textbf{Configuration}}
    & \multicolumn{2}{c}{\textbf{General}}
    & \multicolumn{4}{c}{\textbf{Grounding}} \\
    \cmidrule(lr){2-3}
    \cmidrule(lr){4-7}
    & \textbf{Visual}
    & \textbf{Spatial}
    & \textbf{IoU}
    & \textbf{ACC@0.5}
    & \textbf{ACC@0.7}
    & \textbf{ACC@0.9} \\
    \midrule
    Original
    &\underline{61.8} &37.8  &74.1 &\underline{85.9} &73.3 &\underline{26.5} \\
    w/o T-V Spatial
    &\textbf{61.9} &\textbf{39.9}  &\textbf{74.4} &\textbf{86.0} &\textbf{74.3} &\textbf{28.0}\\
    w/o V-V Spatial
    &\underline{61.8} &\underline{38.4} &73.3 &85.7 &72.1 &23.9 \\
    w/o Both Spatial
    &61.4 &38.1 &73.4 &\underline{85.9} &72.1 &23.4 \\
    \bottomrule
    \end{tabular}
    }
    \label{tab:ilrope}
\end{table}

\section{Proofs}
\subsection{Equivalence of Operator Spatial Neutrality}
\label{app:neutrality_equivalence}

\begin{proof}
Consider a given text token $x$ and any two visual patches $a,b$ within the same image. By Definition~\ref{eq:definition}, cross-modal spatial neutrality requires
\begin{equation}
\mathbf{q}^{\top}
\left[
R(\mathbf{p}_a-\mathbf{p}_x)
-
R(\mathbf{p}_b-\mathbf{p}_x)
\right]
\mathbf{k}
=0,
\quad
\forall\,\mathbf{q},\mathbf{k}\in\mathbb{R}^d.
\end{equation}
Let $\{\mathbf{e}_1,\ldots,\mathbf{e}_d\}$ be the
standard basis of $\mathbb{R}^d$.
Choosing $\mathbf{q}=\mathbf{e}_r$ and
$\mathbf{k}=\mathbf{e}_s$ for arbitrary
$r,s\in\{1,\ldots,d\}$ yields
\begin{equation}
\left[
R(\mathbf{p}_a-\mathbf{p}_x)
-
R(\mathbf{p}_b-\mathbf{p}_x)
\right]_{rs}
=0.
\end{equation}
Since every matrix entry vanishes, we obtain
\begin{equation}
R(\mathbf{p}_a-\mathbf{p}_x)
=
R(\mathbf{p}_b-\mathbf{p}_x).
\end{equation}
Conversely, equality of the rotation operators
immediately implies equality of the corresponding
bilinear forms for arbitrary $\mathbf{q},\mathbf{k}$.
Applying this equivalence to all visual patch pairs
$a,b$ establishes Eq.~\ref{eq:operator_neutrality}.
\end{proof}

\subsection{Phase Condition for Cross-Modal Spatial Neutrality}
\label{app:circle_proof}

\begin{proof}
Let $\mathbf{p}_a'$ and $\mathbf{p}_b'$ denote the projected coordinates of two visual patches under Circle-RoPE. By the block-diagonal structure of
MRoPE, the operator equality in Eq.~\ref{eq:operator_neutrality} holds if and only if the corresponding rotary blocks are equal:
\begin{equation}
\begin{aligned}
&R(\mathbf{p}_a'-\mathbf{p}_x)
=R(\mathbf{p}_b'-\mathbf{p}_x)\\
&\iff
\rho\left(
\theta_\ell
[\mathbf{p}_a'-\mathbf{p}_x]_{a(\ell)}
\right)
=
\rho\left(
\theta_\ell
[\mathbf{p}_b'-\mathbf{p}_x]_{a(\ell)}
\right),
\quad \forall\,\ell.
\end{aligned}
\end{equation}
Since $\rho(\alpha)=\rho(\beta)$ if and only if $\alpha-\beta\in2\pi\mathbb{Z}$, the above condition is equivalent to
\begin{equation}
\theta_\ell
[\mathbf{p}_a'-\mathbf{p}_b']_{a(\ell)}
\in2\pi\mathbb{Z},
\qquad \forall\,\ell,
\end{equation}
which establishes Eq.~\ref{eq:circle_phase}.
\end{proof}

\subsection{Proof of Observation~\ref{eq:theorem}}
\label{app:neutrality_theorem}

\begin{proof}
Consider a given text token $x$ and any two visual patches $a,b$ within the same image. For the shared RoPE operator, the planar rotation
identities
$\rho(\alpha)^\top=\rho(-\alpha)$ and
$\rho(\alpha)\rho(\beta)=\rho(\alpha+\beta)$
imply
\begin{equation}
R(\mathbf{p}_a-\mathbf{p}_x)^\top
R(\mathbf{p}_b-\mathbf{p}_x)=
R(\mathbf{p}_b-\mathbf{p}_a).
\label{eq:app_rotation_relation}
\end{equation}

\textit{($\Rightarrow$) Necessity.}
Suppose cross-modal spatial neutrality holds.
By Eq.~\ref{eq:operator_neutrality},
\begin{equation}
R(\mathbf{p}_a-\mathbf{p}_x)
=
R(\mathbf{p}_b-\mathbf{p}_x),
\quad \forall\,a,b.
\end{equation}
Substituting this equality into
Eq.~\ref{eq:app_rotation_relation} and using
the orthogonality of RoPE rotations yields
\begin{equation}
R(\mathbf{p}_b-\mathbf{p}_a)=
R(\mathbf{p}_a-\mathbf{p}_x)^\top
R(\mathbf{p}_a-\mathbf{p}_x)=I.
\end{equation}
Thus, all intra-visual relative rotations are trivial.

\textit{($\Leftarrow$) Sufficiency.}
Conversely, suppose
$R(\mathbf{p}_b-\mathbf{p}_a)=I$
for all visual patch pairs $a,b$.
By Eq.~\ref{eq:app_rotation_relation},
\begin{equation}
R(\mathbf{p}_a-\mathbf{p}_x)^\top
R(\mathbf{p}_b-\mathbf{p}_x)
=I.
\end{equation}
Multiplying both sides from the left by $R(\mathbf{p}_a-\mathbf{p}_x)$ gives
\begin{equation}
R(\mathbf{p}_b-\mathbf{p}_x)
=
R(\mathbf{p}_a-\mathbf{p}_x),
\quad \forall\,a,b.
\end{equation}
Therefore, Eq.~\ref{eq:operator_neutrality} holds, establishing cross-modal spatial neutrality. The two conditions are thus equivalent.
\end{proof}

\section{Implementation Details}
\label{app:implementation_details}

All controlled training experiments are conducted using
LLaMA-Factory~\cite{zheng2024llamafactory} on four NVIDIA H200 GPUs. We use identical
training hyperparameters for Qwen3-VL-2B and Qwen3-VL-4B
across all spatial configurations. We adopt eager attention
to implement our attention-specific coordinate assignments.
The detailed training settings are summarized in
Table~\ref{tab:training_hyperparams}.

\begin{table}[t]
    \centering
    \caption{Hyperparameter configurations for the two-stage
    controlled training protocol. Batch size denotes the
    effective global batch size.}
    \label{tab:training_hyperparams}
    \resizebox{\columnwidth}{!}{
    \begin{tabular}{lcc}
        \toprule
        \textbf{Configuration} & \textbf{Stage 1} & \textbf{Stage 2} \\
        \midrule
        Trainable Modules & Projector + Merger & All Modules \\
        Max Length & 8,192 & 8,192 \\
        Image Resolution & Dynamic & Dynamic \\
        Image Pixel Range & 1,024--1,048,576 & 1,024--1,048,576 \\
        Attention Backend & Eager & Eager \\
        Optimizer & AdamW & AdamW \\
        Batch Size & 256 & 128 \\
        Learning Rate & $1\times10^{-3}$ & $2\times10^{-5}$ \\
        LR Schedule & Cosine & Cosine \\
        Warmup Ratio & 0.03 & 0.03 \\
        Weight Decay & 0.0 & 0.0 \\
        Max Grad Norm & 1.0 & 1.0 \\
        Epochs & 1 & 1 \\
        Numerical Precision & BF16 & BF16 \\
        DeepSpeed Stage & ZeRO-2 & ZeRO-2 \\
        \bottomrule
    \end{tabular}
    }
\end{table}

\section{More Examples of Attention Map Visualization}
\label{app:attn}
\begin{figure*}[t]
    \begin{minipage}{0.05\linewidth}
        \mbox{}
    \end{minipage}%
    \begin{minipage}{0.9\linewidth}
        \begin{minipage}{0.245\linewidth}
        \centering
        Input: Where is the \textcolor{orange}{cat}?
    \end{minipage}%
    \end{minipage}%
    \par
    \begin{minipage}{0.05\linewidth}
        \mbox{}
    \end{minipage}%
    \begin{minipage}{0.24\linewidth}
    \includegraphics[width=\linewidth]{figure/appendix_attn/cat_on_a_tree.jpg}
    \end{minipage}
    \par
    \rule{\textwidth}{0.5pt}
    \begin{minipage}{\linewidth}
    \begin{minipage}{0.05\linewidth}
        \mbox{}
    \end{minipage}%
    \begin{minipage}{0.9\linewidth}
        \begin{minipage}{0.25\linewidth}
            \centering Original
        \end{minipage}%
        \begin{minipage}{0.25\linewidth}
            \centering w/o T-V
        \end{minipage}%
        \begin{minipage}{0.25\linewidth}
            \centering w/o V-V
        \end{minipage}%
        \begin{minipage}{0.25\linewidth}
            \centering w/o Both
        \end{minipage}%
    \end{minipage}
    \end{minipage}
    \par
    \begin{minipage}{\linewidth}
        \begin{minipage}{0.05\linewidth}
            \centering
            \rotatebox{90}{Shallow}
        \end{minipage}
        \begin{minipage}{0.9\linewidth}
            \includegraphics[width=\linewidth]{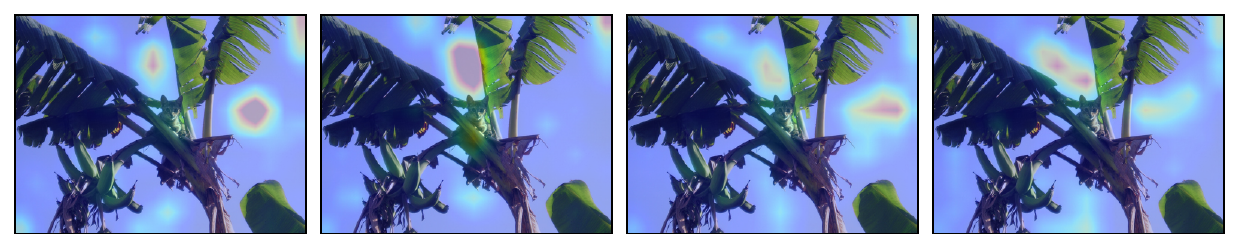}
        \end{minipage}
    \end{minipage}
    \par
    \begin{minipage}{\linewidth}
        \begin{minipage}{0.05\linewidth}
            \centering
            \rotatebox{90}{Middle}
        \end{minipage}
        \begin{minipage}{0.9\linewidth}
            \includegraphics[width=\linewidth]{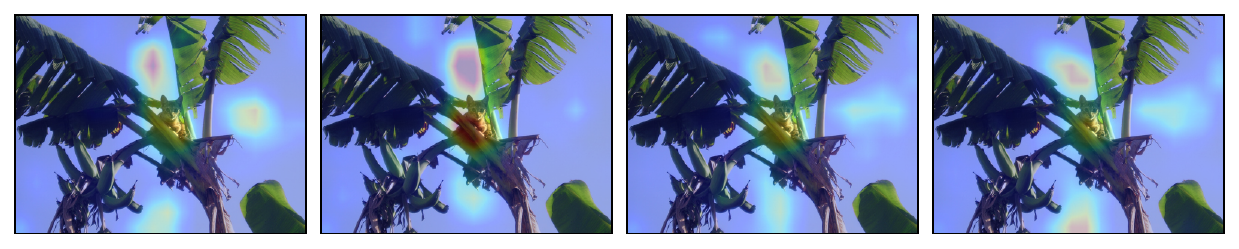}
        \end{minipage}
    \end{minipage}
    \par
    \begin{minipage}{\linewidth}
        \begin{minipage}{0.05\linewidth}
            \centering
            \rotatebox{90}{Deep}
        \end{minipage}
        \begin{minipage}{0.9\linewidth}
            \includegraphics[width=\linewidth]{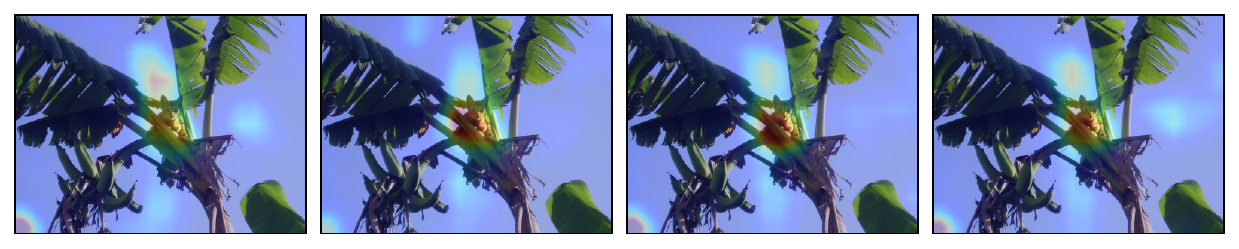}
        \end{minipage}
    \end{minipage}
    \par
    \rule{\textwidth}{0.5pt}
    \setlength{\fboxsep}{2pt}
    \caption{
    Visualization of attention maps for Qwen3-VL-4B
    using the MRoPE-I under four spatial configurations.
    The orange-highlighted word \textcolor{orange}{cat} indicates the query text token used to compute attention over visual patches.
    Columns correspond to Original, w/o T-V Spatial, w/o V-V Spatial,
    and w/o Both Spatial. Rows represent attention averaged over shallow (layers 0-11), middle (layers 12-23), and deep (layers 24-35) layers. Warmer colors indicate higher relative attention intensity.
    }
    \label{fig:appendix_mropei_attn_img1}
\end{figure*}

\begin{figure*}[t]
    \begin{minipage}{0.05\linewidth}
        \mbox{}
    \end{minipage}%
    \begin{minipage}{0.9\linewidth}
        \begin{minipage}{0.245\linewidth}
        \centering
        Input: Where is the \textcolor{orange}{cat}?
    \end{minipage}%
    \end{minipage}%
    \par
    \begin{minipage}{0.05\linewidth}
        \mbox{}
    \end{minipage}%
    \begin{minipage}{0.24\linewidth}
    \includegraphics[width=\linewidth]{figure/appendix_attn/cat_on_a_tree.jpg}
    \end{minipage}
    \par
    \rule{\textwidth}{0.5pt}
    \begin{minipage}{\linewidth}
    \begin{minipage}{0.05\linewidth}
        \mbox{}
    \end{minipage}%
    \begin{minipage}{0.9\linewidth}
        \begin{minipage}{0.25\linewidth}
            \centering Original
        \end{minipage}%
        \begin{minipage}{0.25\linewidth}
            \centering w/o T-V
        \end{minipage}%
        \begin{minipage}{0.25\linewidth}
            \centering w/o V-V
        \end{minipage}%
        \begin{minipage}{0.25\linewidth}
            \centering w/o Both
        \end{minipage}%
    \end{minipage}
    \end{minipage}
    \par
    \begin{minipage}{\linewidth}
        \begin{minipage}{0.05\linewidth}
            \centering
            \rotatebox{90}{Shallow}
        \end{minipage}
        \begin{minipage}{0.9\linewidth}
            \includegraphics[width=\linewidth]{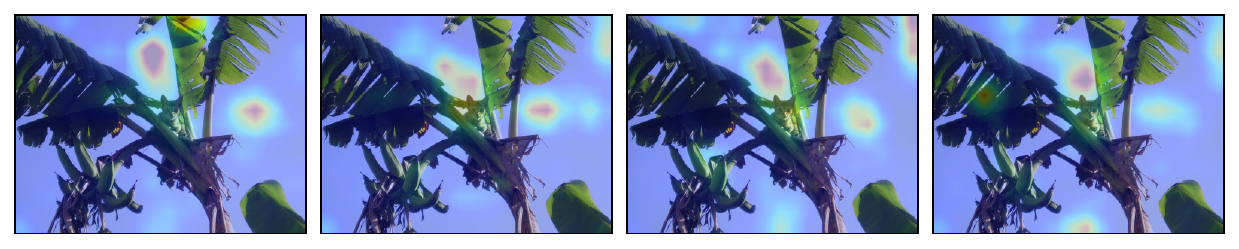}
        \end{minipage}
    \end{minipage}
    \par
    \begin{minipage}{\linewidth}
        \begin{minipage}{0.05\linewidth}
            \centering
            \rotatebox{90}{Middle}
        \end{minipage}
        \begin{minipage}{0.9\linewidth}
            \includegraphics[width=\linewidth]{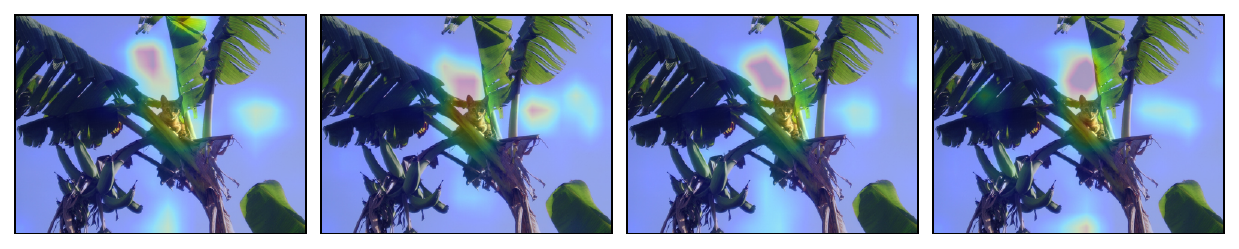}
        \end{minipage}
    \end{minipage}
    \par
    \begin{minipage}{\linewidth}
        \begin{minipage}{0.05\linewidth}
            \centering
            \rotatebox{90}{Deep}
        \end{minipage}
        \begin{minipage}{0.9\linewidth}
            \includegraphics[width=\linewidth]{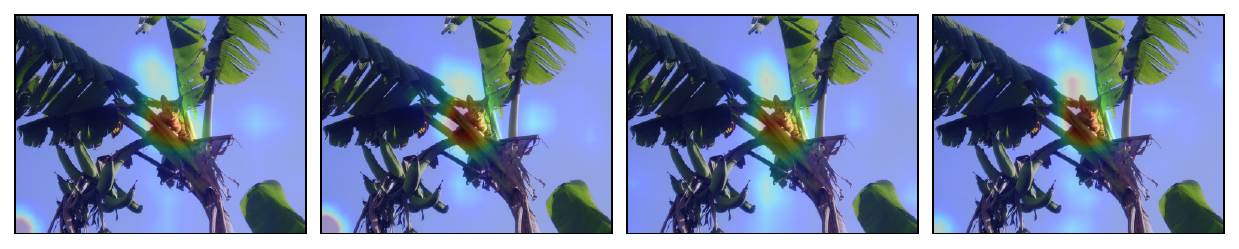}
        \end{minipage}
    \end{minipage}
    \par
    \rule{\textwidth}{0.5pt}
    \setlength{\fboxsep}{2pt}
    \caption{Visualization of attention maps for Qwen3-VL-4B
    using the VideoRoPE under four spatial configurations.
    The orange-highlighted word \textcolor{orange}{cat} indicates the query text token used to compute attention over visual patches.
    Columns correspond to Original, w/o T-V Spatial, w/o V-V Spatial,
    and w/o Both Spatial. Rows represent attention averaged over shallow (layers 0-11), middle (layers 12-23), and deep (layers 24-35) layers. Warmer colors indicate higher relative attention intensity.}
    \label{fig:appendix_videorope_attn_img1}
\end{figure*}

\begin{figure*}[t]
    \begin{minipage}{0.05\linewidth}
        \mbox{}
    \end{minipage}%
    \begin{minipage}{0.9\linewidth}
        \begin{minipage}{0.245\linewidth}
        \centering
        Input: Where is the \textcolor{orange}{cat}?
    \end{minipage}%
    \end{minipage}%
    \par
    \begin{minipage}{0.05\linewidth}
        \mbox{}
    \end{minipage}%
    \begin{minipage}{0.24\linewidth}
    \includegraphics[width=\linewidth]{figure/appendix_attn/cat_on_a_tree.jpg}
    \end{minipage}
    \par
    \rule{\textwidth}{0.5pt}
    \begin{minipage}{\linewidth}
    \begin{minipage}{0.05\linewidth}
        \mbox{}
    \end{minipage}%
    \begin{minipage}{0.9\linewidth}
        \begin{minipage}{0.25\linewidth}
            \centering Original
        \end{minipage}%
        \begin{minipage}{0.25\linewidth}
            \centering w/o T-V
        \end{minipage}%
        \begin{minipage}{0.25\linewidth}
            \centering w/o V-V
        \end{minipage}%
        \begin{minipage}{0.25\linewidth}
            \centering w/o Both
        \end{minipage}%
    \end{minipage}
    \end{minipage}
    \par
    \begin{minipage}{\linewidth}
        \begin{minipage}{0.05\linewidth}
            \centering
            \rotatebox{90}{Shallow}
        \end{minipage}
        \begin{minipage}{0.9\linewidth}
            \includegraphics[width=\linewidth]{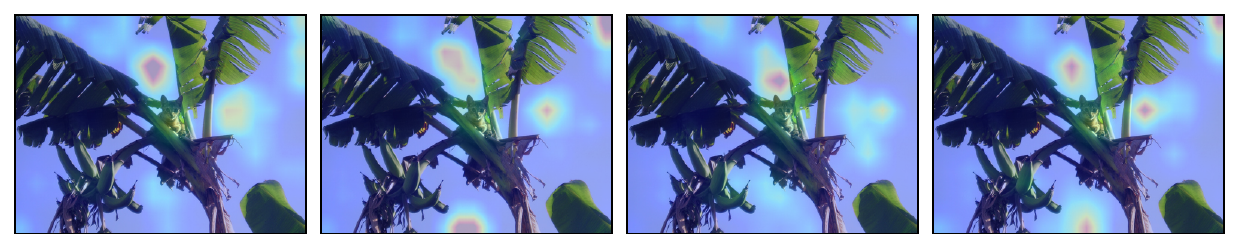}
        \end{minipage}
    \end{minipage}
    \par
    \begin{minipage}{\linewidth}
        \begin{minipage}{0.05\linewidth}
            \centering
            \rotatebox{90}{Middle}
        \end{minipage}
        \begin{minipage}{0.9\linewidth}
            \includegraphics[width=\linewidth]{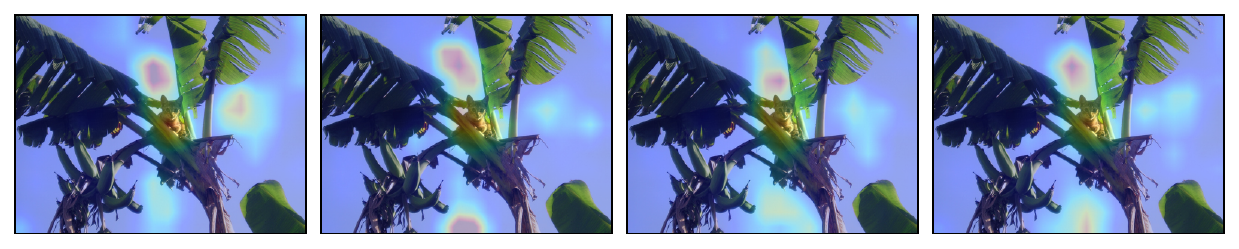}
        \end{minipage}
    \end{minipage}
    \par
    \begin{minipage}{\linewidth}
        \begin{minipage}{0.05\linewidth}
            \centering
            \rotatebox{90}{Deep}
        \end{minipage}
        \begin{minipage}{0.9\linewidth}
            \includegraphics[width=\linewidth]{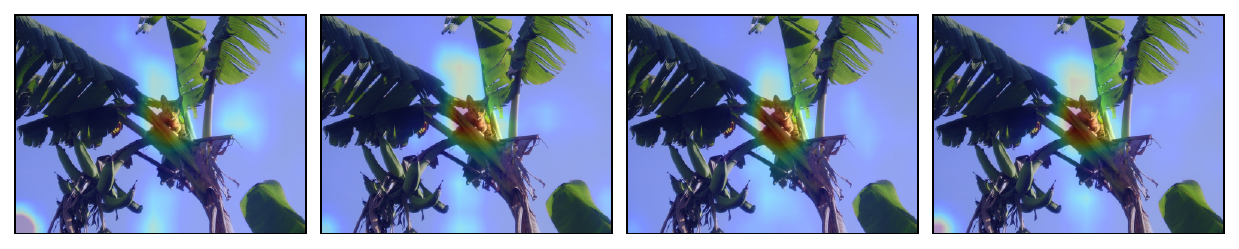}
        \end{minipage}
    \end{minipage}
    \par
    \rule{\textwidth}{0.5pt}
    \setlength{\fboxsep}{2pt}
    \caption{Visualization of attention maps for Qwen3-VL-4B
    using the HoPE under four spatial configurations.
    The orange-highlighted word \textcolor{orange}{cat} indicates the query text token used to compute attention over visual patches.
    Columns correspond to Original, w/o T-V Spatial, w/o V-V Spatial,
    and w/o Both Spatial. Rows represent attention averaged over shallow (layers 0-11), middle (layers 12-23), and deep (layers 24-35) layers. Warmer colors indicate higher relative attention intensity.}
    \label{fig:appendix_hope_attn_img1}
\end{figure*}

\begin{figure*}[t]
    \begin{minipage}{0.05\linewidth}
        \mbox{}
    \end{minipage}%
    \begin{minipage}{0.9\linewidth}
        \begin{minipage}{0.245\linewidth}
        \centering
        Input: How many traffic lights are \textcolor{orange}{green}?
    \end{minipage}%
    \end{minipage}%
    \par
    \begin{minipage}{0.05\linewidth}
        \mbox{}
    \end{minipage}%
    \begin{minipage}{0.24\linewidth}
    \includegraphics[width=\linewidth]{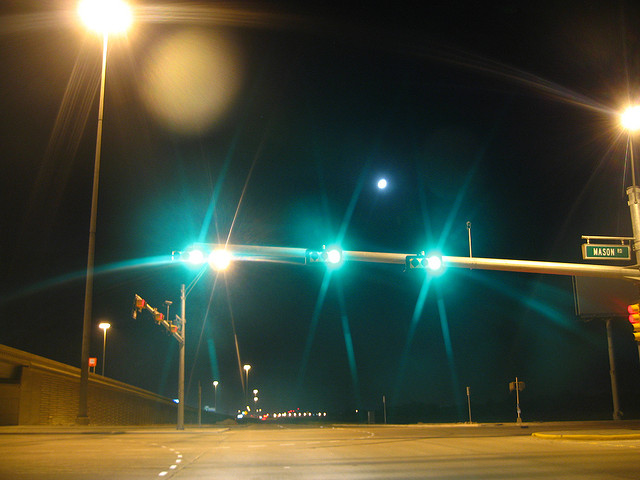}
    \end{minipage}
    \par
    \rule{\textwidth}{0.5pt}
    \begin{minipage}{\linewidth}
    \begin{minipage}{0.05\linewidth}
        \mbox{}
    \end{minipage}%
    \begin{minipage}{0.9\linewidth}
        \begin{minipage}{0.25\linewidth}
            \centering Original
        \end{minipage}%
        \begin{minipage}{0.25\linewidth}
            \centering w/o T-V
        \end{minipage}%
        \begin{minipage}{0.25\linewidth}
            \centering w/o V-V
        \end{minipage}%
        \begin{minipage}{0.25\linewidth}
            \centering w/o Both
        \end{minipage}%
    \end{minipage}
    \end{minipage}
    \par
    \begin{minipage}{\linewidth}
        \begin{minipage}{0.05\linewidth}
            \centering
            \rotatebox{90}{Shallow}
        \end{minipage}
        \begin{minipage}{0.9\linewidth}
            \includegraphics[width=\linewidth]{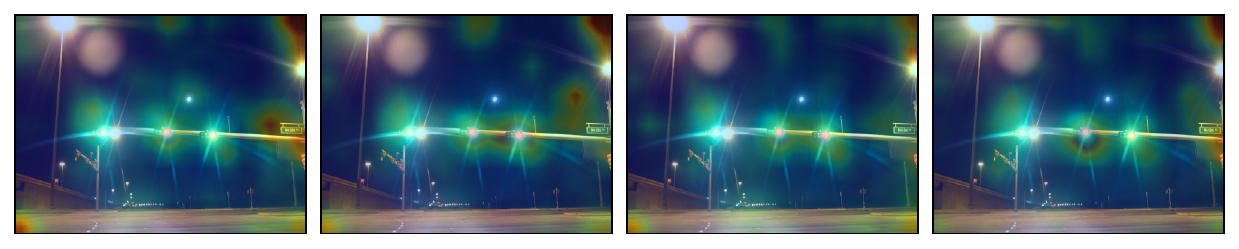}
        \end{minipage}
    \end{minipage}
    \par
    \begin{minipage}{\linewidth}
        \begin{minipage}{0.05\linewidth}
            \centering
            \rotatebox{90}{Middle}
        \end{minipage}
        \begin{minipage}{0.9\linewidth}
            \includegraphics[width=\linewidth]{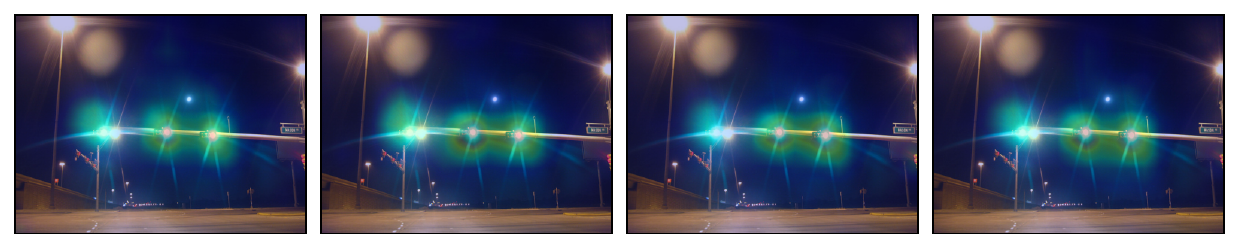}
        \end{minipage}
    \end{minipage}
    \par
    \begin{minipage}{\linewidth}
        \begin{minipage}{0.05\linewidth}
            \centering
            \rotatebox{90}{Deep}
        \end{minipage}
        \begin{minipage}{0.9\linewidth}
            \includegraphics[width=\linewidth]{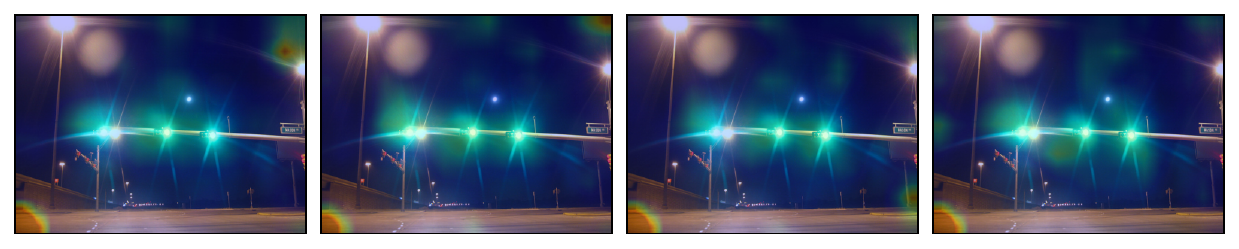}
        \end{minipage}
    \end{minipage}
    \par
    \rule{\textwidth}{0.5pt}
    \setlength{\fboxsep}{2pt}
    \caption{Visualization of attention maps for Qwen3-VL-4B
    using the MRoPE-I under four spatial configurations.
    The orange-highlighted word \textcolor{orange}{green} indicates the query text token used to compute attention over visual patches. Columns correspond to Original, w/o T-V Spatial, w/o V-V Spatial, and w/o Both Spatial. Rows represent attention averaged over shallow (layers 0-11), middle (layers 12-23), and deep (layers 24-35) layers. Warmer colors indicate higher relative attention intensity.}
    \label{fig:appendix_mrope_attn_img2}
\end{figure*}

\begin{figure*}[t]
    \begin{minipage}{0.05\linewidth}
        \mbox{}
    \end{minipage}%
    \begin{minipage}{0.9\linewidth}
        \begin{minipage}{0.245\linewidth}
        \centering
        Input: How many traffic lights are \textcolor{orange}{green}?
    \end{minipage}%
    \end{minipage}%
    \par
    \begin{minipage}{0.05\linewidth}
        \mbox{}
    \end{minipage}%
    \begin{minipage}{0.24\linewidth}
    \includegraphics[width=\linewidth]{figure/appendix_attn/lights.jpg}
    \end{minipage}
    \par
    \rule{\textwidth}{0.5pt}
    \begin{minipage}{\linewidth}
    \begin{minipage}{0.05\linewidth}
        \mbox{}
    \end{minipage}%
    \begin{minipage}{0.9\linewidth}
        \begin{minipage}{0.25\linewidth}
            \centering Original
        \end{minipage}%
        \begin{minipage}{0.25\linewidth}
            \centering w/o T-V
        \end{minipage}%
        \begin{minipage}{0.25\linewidth}
            \centering w/o V-V
        \end{minipage}%
        \begin{minipage}{0.25\linewidth}
            \centering w/o Both
        \end{minipage}%
    \end{minipage}
    \end{minipage}
    \par
    \begin{minipage}{\linewidth}
        \begin{minipage}{0.05\linewidth}
            \centering
            \rotatebox{90}{Shallow}
        \end{minipage}
        \begin{minipage}{0.9\linewidth}
            \includegraphics[width=\linewidth]{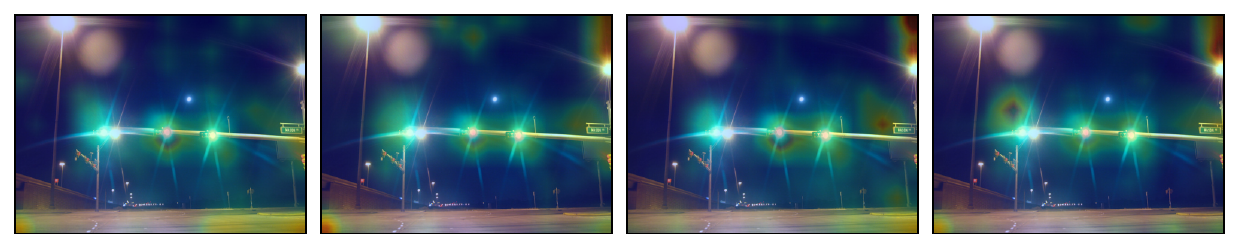}
        \end{minipage}
    \end{minipage}
    \par
    \begin{minipage}{\linewidth}
        \begin{minipage}{0.05\linewidth}
            \centering
            \rotatebox{90}{Middle}
        \end{minipage}
        \begin{minipage}{0.9\linewidth}
            \includegraphics[width=\linewidth]{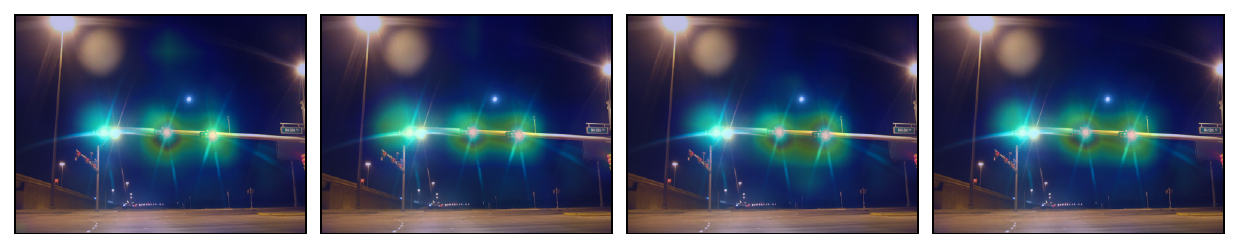}
        \end{minipage}
    \end{minipage}
    \par
    \begin{minipage}{\linewidth}
        \begin{minipage}{0.05\linewidth}
            \centering
            \rotatebox{90}{Deep}
        \end{minipage}
        \begin{minipage}{0.9\linewidth}
            \includegraphics[width=\linewidth]{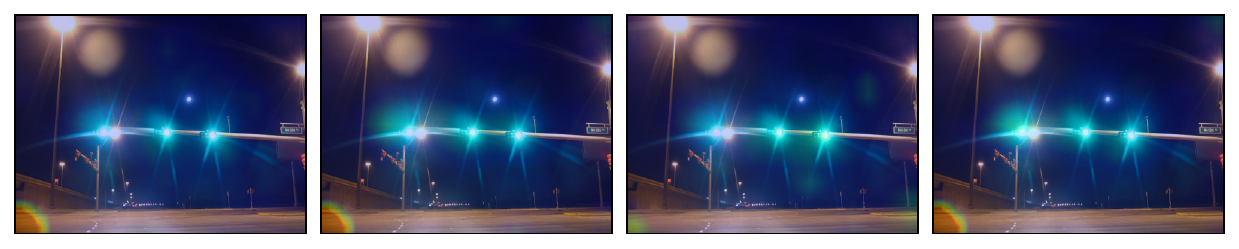}
        \end{minipage}
    \end{minipage}
    \par
    \rule{\textwidth}{0.5pt}
    \setlength{\fboxsep}{2pt}
    \caption{Visualization of attention maps for Qwen3-VL-4B
    using the VideoRoPE under four spatial configurations.
    The orange-highlighted word \textcolor{orange}{green} indicates the query text token used to compute attention over visual patches. Columns correspond to Original, w/o T-V Spatial, w/o V-V Spatial, and w/o Both Spatial. Rows represent attention averaged over shallow (layers 0-11), middle (layers 12-23), and deep (layers 24-35) layers. Warmer colors indicate higher relative attention intensity.}
    \label{fig:appendix_videorope_attn_img2}
\end{figure*}

\begin{figure*}[t]
    \begin{minipage}{0.05\linewidth}
        \mbox{}
    \end{minipage}%
    \begin{minipage}{0.9\linewidth}
        \begin{minipage}{0.245\linewidth}
        \centering
        Input: How many traffic lights are \textcolor{orange}{green}?
    \end{minipage}%
    \end{minipage}%
    \par
    \begin{minipage}{0.05\linewidth}
        \mbox{}
    \end{minipage}%
    \begin{minipage}{0.24\linewidth}
    \includegraphics[width=\linewidth]{figure/appendix_attn/lights.jpg}
    \end{minipage}
    \par
    \rule{\textwidth}{0.5pt}
    \begin{minipage}{\linewidth}
    \begin{minipage}{0.05\linewidth}
        \mbox{}
    \end{minipage}%
    \begin{minipage}{0.9\linewidth}
        \begin{minipage}{0.25\linewidth}
            \centering Original
        \end{minipage}%
        \begin{minipage}{0.25\linewidth}
            \centering w/o T-V
        \end{minipage}%
        \begin{minipage}{0.25\linewidth}
            \centering w/o V-V
        \end{minipage}%
        \begin{minipage}{0.25\linewidth}
            \centering w/o Both
        \end{minipage}%
    \end{minipage}
    \end{minipage}
    \par
    \begin{minipage}{\linewidth}
        \begin{minipage}{0.05\linewidth}
            \centering
            \rotatebox{90}{Shallow}
        \end{minipage}
        \begin{minipage}{0.9\linewidth}
            \includegraphics[width=\linewidth]{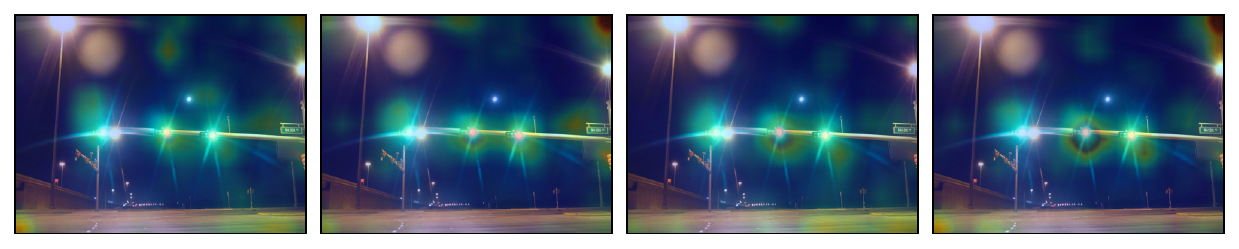}
        \end{minipage}
    \end{minipage}
    \par
    \begin{minipage}{\linewidth}
        \begin{minipage}{0.05\linewidth}
            \centering
            \rotatebox{90}{Middle}
        \end{minipage}
        \begin{minipage}{0.9\linewidth}
            \includegraphics[width=\linewidth]{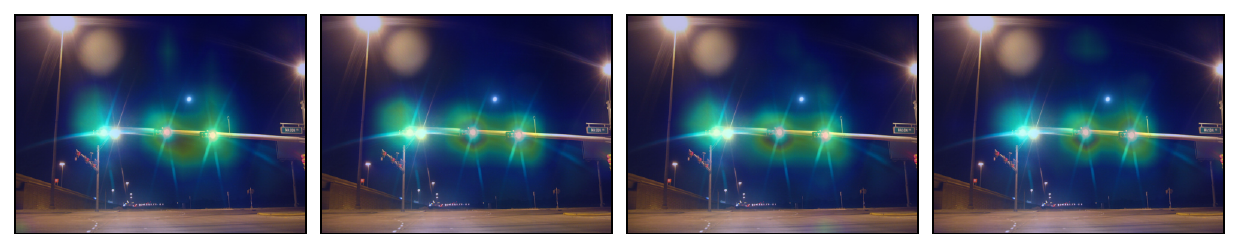}
        \end{minipage}
    \end{minipage}
    \par
    \begin{minipage}{\linewidth}
        \begin{minipage}{0.05\linewidth}
            \centering
            \rotatebox{90}{Deep}
        \end{minipage}
        \begin{minipage}{0.9\linewidth}
            \includegraphics[width=\linewidth]{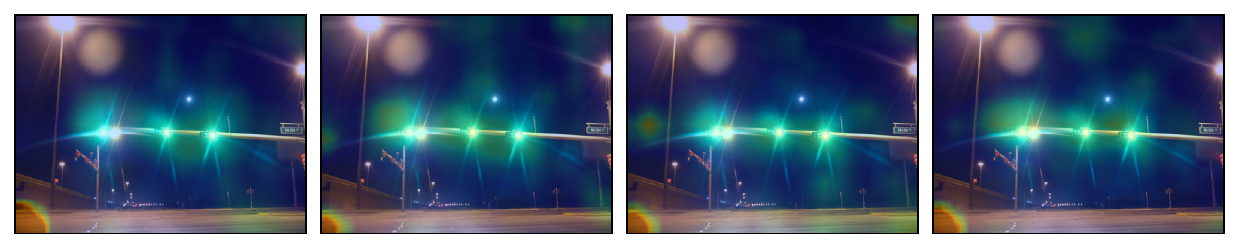}
        \end{minipage}
    \end{minipage}
    \par
    \rule{\textwidth}{0.5pt}
    \setlength{\fboxsep}{2pt}
    \caption{Visualization of attention maps for Qwen3-VL-4B
    using the HoPE under four spatial configurations.
    The orange-highlighted word \textcolor{orange}{green} indicates the query text token used to compute attention over visual patches. Columns correspond to Original, w/o T-V Spatial, w/o V-V Spatial, and w/o Both Spatial. Rows represent attention averaged over shallow (layers 0-11), middle (layers 12-23), and deep (layers 24-35) layers. Warmer colors indicate higher relative attention intensity.}
    \label{fig:appendix_hope_attn_img2}
\end{figure*}

More visualizations of attention maps are illustrated from Fig.~\ref{fig:appendix_mropei_attn_img1} to Fig.~\ref{fig:appendix_hope_attn_img2}.

\newpage

\end{document}